\documentclass[acmtog,nonacm]{acmart} 
\acmSubmissionID{0438}

\usepackage{booktabs} 
\usepackage{pgfplots}
\usepackage[ruled]{algorithm2e} 

\SetAlFnt{\small}
\SetAlCapFnt{\small}
\SetAlCapNameFnt{\small}
\SetAlCapHSkip{0pt}
\graphicspath{{images/}{images/out/}}

\newcommand{\neat}{NeAT\xspace}

\newcommand{\vvec}[1]{\mathbf{#1}}   
\newcommand{\mmat}[1]{\mathbf{\uppercase{#1}}}  

\newcommand{\loss}{\mathcal{L}}
\newcommand{\err}{E}

\newcommand{\img}{I}          
\newcommand{\bgimg}{\img_0}   
\newcommand{\oimg}{\hat\img}  
\newcommand{\obgimg}{\hat\img_0}  
\newcommand{\pix}{p}        

\newcommand{\density}{\sigma} 
\newcommand{\mlp}{\Phi}       

\newcommand{\ray}{r}
\newcommand{\ttt}{t}          
\newcommand{\ts}{{t_n}}       
\newcommand{\te}{{t_f}}       
\newcommand{\x}{\vvec{x}}     
\newcommand{\feat}{f}         

\begin{document}
\title{NeAT: Neural Adaptive Tomography}

\author{Darius R\"uckert}
\email{darius.rueckert@fau.de}
\affiliation{%
  \institution{KAUST and University of Erlangen-Nuremberg}
  \city{Erlangen}
  \country{Germany}
}

\author{Yuanhao Wang}
\author{Rui Li}
\author{Ramzi Idoughi}
\author{Wolfgang Heidrich}
\email{wolfgang.heidrich@kaust.edu.sa}
\affiliation{%
  \institution{KAUST}
  \city{Thuwal}
  \country{Saudi Arabia}
}
\renewcommand\shortauthors{R\"uckert et al.}

\begin{teaserfigure}
	\includegraphics[width=\linewidth]{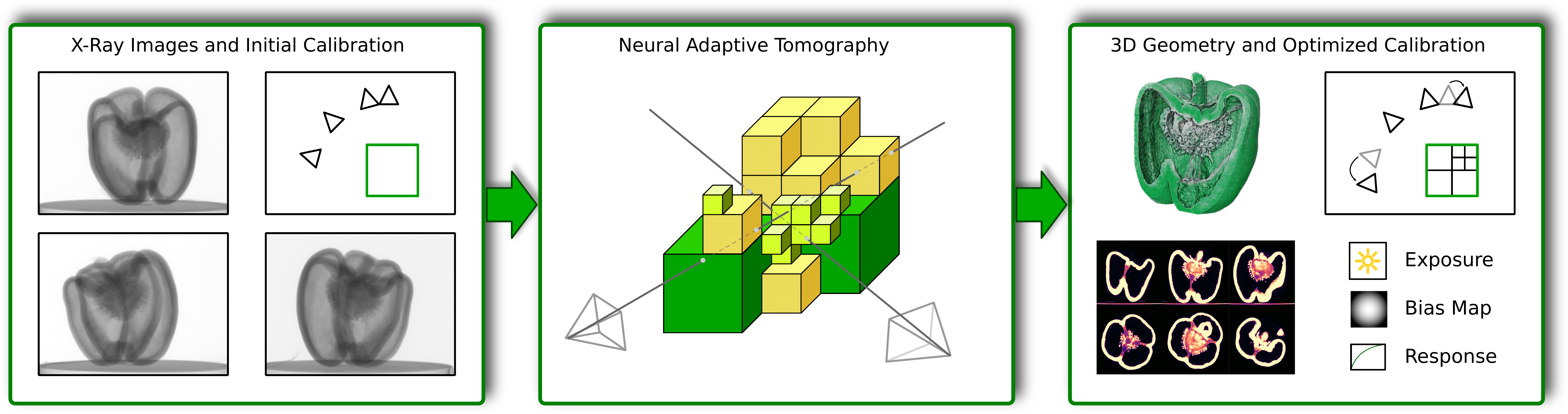}
	\caption{Neural Adaptive Tomography uses a hybrid
          explicit-implicit neural representation for tomographic
          image reconstruction. Left: The input is a set of X-ray
          images, typically with an ill-posed geometric configuration
          (sparse views or limited angular coverage). Center: \neat
          represents the scene as an octree with neural features in
          each leaf node. This representation lends itself to an
          efficient differentiable rendering algorithm, presented in
          this paper. Right: Through neural rendering \neat can
          reconstruct the 3D geometry even for ill-posed
          configurations, while simultaneously performing geometric
          and radiometric self-calibration.}
	\label{fig:teaser}
\end{teaserfigure}



\begin{abstract}
  In this paper, we present Neural Adaptive Tomography (\neat), the
  first adaptive, hierarchical neural rendering pipeline for
  multi-view inverse rendering. Through a combination of neural
  features with an adaptive explicit representation, we achieve
  reconstruction times far superior to existing neural inverse
  rendering methods. The adaptive explicit representation improves
  efficiency by facilitating empty space culling and concentrating
  samples in complex regions, while the neural features act as a
  neural regularizer for the 3D reconstruction.

  The \neat framework is designed specifically for the tomographic
  setting, which consists only of semi-transparent volumetric scenes
  instead of opaque objects. In this setting, \neat outperforms the
  quality of existing optimization-based tomography solvers while being
  substantially faster.
\end{abstract}


%
%


\maketitle

\section{Introduction}

Computed Tomography (CT) is an important scientific imaging modality
in a wide range of fields, from medical imaging to material
science. While most CT imaging is performed with X-rays due to their
ability to penetrate a wide range of
materials~\cite{kak2001principles}, there have also been a number of
works utilizing visible light, especially in the visual computing
community (e.g.~\cite{hasinoff2007photo, atcheson2008time,
  gregson2012stochastic, eckert2019scalarflow, zang2020tomofluid}).

The tomographic reconstruction problem is the task of estimating the
3D structure of a sample from its 2D projections. This task is well-posed
under certain conditions, such as a sufficiently large number of
projections/views, good angular coverage of these views, and low
noise. In this situation, transform-based methods like filtered
backprojection~\cite{feldkamp1984practical} provide a fast and
accurate reconstruction. Unfortunately, these methods no longer
produce satisfactory results if the above conditions are violated
(small number of views, poor angular distribution, or high noise). For
these types of difficult settings, a range of iterative
optimization-based methods have been developed in recent years
(e.g.~\cite{sidky2008image, huang2013iterative, huang2018scale,
  zang2018super, xu2020limited}. These new methods greatly expand the
envelope of feasible tomographic reconstruction problems, albeit at a
significantly increased computational cost.

In parallel to this development, both neural rendering
(e.g.~\cite{liu2020neural, reiser2021kilonerf, garbin2021fastnerf})
and differentiable rendering in general
(e.g.~\cite{nimier2019mitsuba}) have recently garnered a lot of
interest in visual computing. In particular, Neural Radiance Fields
(NeRF)~\cite{mildenhall2020nerf}, and related inverse rendering
frameworks have been at the focus of attention due to their ability to
provide superior reconstructions of everyday scenes with opaque
objects. Similar concepts have also already been applied to the
tomographic reconstruction problem~\cite{zang2021intratomo,
  sun2021coil}. However, all the existing neural inverse rendering
frameworks suffer from very long computing times. This is also true
for the tomographic methods, despite operating only on 2D slice geometry,
which has limited their applicability to high-resolution datasets
and to the full 3D cone beam data we use in this work.


In this paper, we present Neural Adaptive Tomography (\neat), the
first adaptive, hierarchical neural rendering pipeline for multi-view
inverse rendering. Through a combination of neural features with an
adaptive explicit representation, we achieve reconstruction times far
superior to existing neural inverse rendering methods. The adaptive
explicit representation improves efficiency by facilitating empty
space culling and concentrating samples in complex regions, while the
neural features act as a neural regularizer for the 3D reconstruction.

\neat is specifically tuned towards tomographic reconstruction
problems, where samples are widely scattered throughout the volume,
while many existing systems like NeRF~\cite{mildenhall2020nerf} rely
on a strong concentration of samples near opaque surfaces. In this
tomographic setting, we demonstrate that the purely explicit
hierarchical representation of \neat outperforms both purely implicit
as well as hybrid explicit-implicit representations akin to
ACORN~\cite{martel2021acorn} in terms of both quality and compute
time. Furthermore, \neat shows improved reconstruction quality
compared to state-of-the-art tomographic reconstruction methods, while
matching their performance.

In summary, the main contributions of our work are:
\begin{itemize}
\item An adaptive, hierarchical neural rendering pipeline based on
  an explicit octree representation with neural features.
\item A differentiable physical sensor model for x-ray imaging that can be optimized during the reconstruction.
\item An efficient open-source implementation that can be readily used on new datasets.
\item An extensive evaluation of our proposed framework on different challenging tomographic reconstruction 
(sparse-view, limited angle, and noisy projections) of both synthetic and real data.
\end{itemize}

\begin{figure*}
	\includegraphics[width=\linewidth]{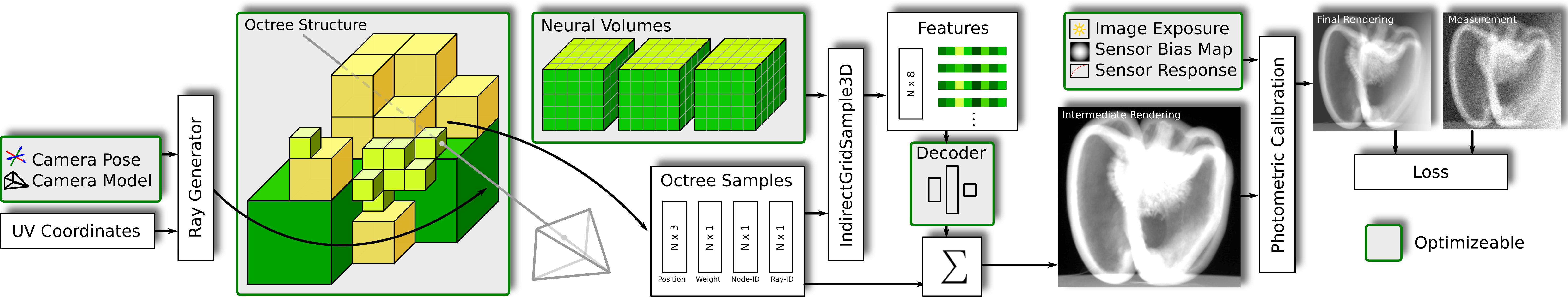}
	\caption{
		Overview of our adaptive neural rendering pipeline for tomographic reconstruction.
		To render a single pixel, we generate the corresponding ray, compute the ray-octree intersection, sample the neural volume, decode the neural features, and integrate them by a weighted sum.
		The estimated pixel value is then passed through a photometric calibration module resulting in the final pixel intensity.
		All elements in green boxes are optimized during the reconstruction.
		This includes the geometric and photometric calibration, the octree structure, the neural features volumes, and the neural decoder network.
	}
	\label{fig:pipeline_overview}
\end{figure*}

\section{Related Work}

\subsection{Classical Computed Tomography}

Computed tomography is a well-established technique used for imaging the internal structures of a scanned object. It has applications in many domains, such as medicine and biology~\cite{van2001computer, kiljunen2015dental, rawson2020x, piovesan2021x}, material science~\cite{brisard2020multiscale, vasarhelyi2020microcomputed}, and fluid dynamics~\cite{hasinoff2007photo, atcheson2008time, gregson2012stochastic, eckert2019scalarflow, zang2020tomofluid}.

In all CT modalities, multiple projection images (sinogram) are captured from different directions. Then, reconstruction algorithms are applied to retrieve a 3D representation of the scanned object from the set of acquired projections. Several algorithm families have been deployed for tomographic reconstruction. Analytic methods based on the Radon transform and its inverse, such as filtered back projection (FBP) and its 3D cone-beam variant FDK (Feldkamp, Davis, and Kress)~\cite{feldkamp1984practical}, are the most used in commercial CT devices~\cite{pan2009commercial}. These methods are fast, and accurate when a large number of uniformly sampled projections is available. However, in many situations, the number of acquired projections is low for a variety of reasons, such as the reduction of the X-ray dose~\cite{gao2014low}, the deformation of the sample~\cite{zang2018space, zang2019warp}, or its inaccessibility from some directions~\cite{du2021study}. For such scenarios, iterative reconstruction approaches have been proposed to solve a discrete formulation of the ill-posed tomography problem. The main interest of these techniques is the possibility to incorporate regularization terms like total variation in an optimization framework~\cite{sidky2008image, huang2013iterative, kisner2012model, huang2018scale, zang2018super, xu2020limited, abujbara2021non}. The hyper-parameter tuning and the high computational requirements are the main downsides of these approaches.

\subsection{Learning-based Computed Tomography}

Recently, learning-based methods have been emerging as an alternative
to optimization-based reconstruction. Most of the initial proposed
approaches apply neural networks either as a pre-processing or a
post-processing step for traditional reconstruction methods to improve
the reconstruction quality. The pre-processing networks improve the
conditioning of the inverse problem by in-painting the
projections~\cite{anirudh2018lose, ghani2018deep, yoo2019sinogram,
  tang2019projection}; while the post-processing networks correct and
denoise the reconstructed volume~\cite{pelt2018improving,
  lucas2018using, liu2020tomogan}. A third strategy consists of using
a network with a differentiable forward model in order to learn a
reconstruction operator~\cite{adler2018learned, chen2018learn,
  kang2018deep, he2020radon}. These approaches achieve high quality
results on data similar to that used for the training. They do,
however, suffer from a substantial lack of generalization when applied
to unseen data.

To overcome this limitation, recent studies introduce the Deep Image
Prior (DIP)~\cite{baguer2020computed, barutcu2021limited} combined
with classical regularization to constraint the reconstruction
problem. On the other hand, some works proposed new approaches based
on an implicit neural representation~\cite{zang2021intratomo,
sun2021coil} to handle the tomography reconstruction in a
self-supervised learning-based fashion. In such methods, a Multi-Layer
Perceptron (MLP) network is used to represent a density field of the
scanned object as a function of the input coordinates. This network is
then learned from the captured projections. This representation offers
an improved flexibility to generate synthetic projections at any
desired resolution. This approach outperforms other existing
techniques in terms of reconstruction quality. However, they are
memory hungry and require a considerable learning time in the range of
hours despite operating only on 2D slices based on parallel beam
data. They are therefore not suitable for full 3D cone beam
reconstruction. In the current paper, we propose an adaptive neural
rendering framework to overcome these limitations and achieve high
quality reconstructions of full 3D cone-beam data in a matter of
minutes.

%
%
%
%

\subsection{Implicit Neural Representations}
In most tomography applications, an explicit, regular voxel grid is
the representation of choice due to the simplicity of the operators.
In computer graphics and computer vision, {\em coordinate-based neural
networks}, also known as {\em implicit neural representations}, have
recently emerged as an alternative.
These consist of a neural network, typically a MLP, to learn functions 
that map spatial coordinates to some physical properties field (e.g. 
occupancy, density, color etc.). The main advantage of this 
representation is that the represented signal
or field is implicitly defined for any given coordinate. In other
words, this representation is continuous, in contrast of the
discretized voxel grids. In the last two years, these coordinate-based
networks have been successfully applied for modeling both static and
dynamic 3D scenes and shapes~\cite{park2019deepsdf,
  sitzmann2020implicit, martin2021nerf, du2021neural, xian2021space},
synthesizing novel views~\cite{eslami2018neural, sitzmann2019scene,
  mildenhall2020nerf, niemeyer2020differentiable, schwarz2020graf,
  chan2021pi}, synthesizing texture~\cite{oechsle2019texture,
  saito2019pifu, chibane2020implicit}, estimating
poses~\cite{yen2020inerf, su2021nerf, wang2021nerf}, and for
relighting and material editing~\cite{boss2021nerd,
  srinivasan2021nerv, xiang2021neutex, zhang2021physg}. In addition to
a huge learning time, coordinate based networks suffer also from a
slow rending speed when switching to a 3D voxel grids. Indeed, the
network has to be evaluated for each single voxel, instead of querying
directly a data structure.

\subsection{Improving Neural Rendering}
Several techniques have been proposed to speed up the volumetric
rendering of coordinate-based neural networks. In Neural Sparse Voxel
Fields (NSVF) approach~\cite{liu2020neural}, the scene is organized
into a sparse voxel octree, which is dynamically updated during the
learning process. During the rendering, empty spaces are skipped, and an
early rays termination are enforced. In kiloNeRF
approach~\cite{reiser2021kilonerf}, the standard NeRF network is
factorized into a 3D grid of smaller MLPs, in order to quicken the
rendering process. The AutoInt technique~\cite{lindell2021autoint} is
based on a network that learns directly the volume integral along a
ray, which makes the rendering step faster in comparison to NeRF
network. FastNeRF~\cite{garbin2021fastnerf} uses caching to have a
faster rendering. The standard NeRF network is split into two MLPs: a
position-dependent network that generates a vector of deep radiance
map, while the second network outputs the corresponding weights for a
given ray direction. Yu et al.~\shortcite{yu2021plenoctrees} proposes
a modified version of NeRF network to predict a volume density and
spherical harmonic weights, which are stored in a "PlenOctree"
structure. This octree structure is then fine-tuned using a rendering
loss to improve its quality. This approach allows a real-time
rendering, however, the training step is still slow.

In parallel to the neural rendering approaches, researchers have also
worked on simply using neural networks to represent existing images
and volumes, without first solving an inverse problem. In the ACORN
approach~\cite{martel2021acorn}, the authors introduce a hybrid
implicit-explicit coordinate neural representation. The learning
process is accelerated through a multi-scale network architecture,
which is optimized during the training in a coarse-to-fine scheme.

\neat is somewhat inspired by all these approaches, but it is the
first to solve a scene reconstruction problem by {\em directly
training a hierarchical neural representation}. We show higher quality
results at drastically improved training times.

%
%
%


\section{Method}

We represent the 3D scene as a sparse octree, where each leaf-node can be empty or contain a uniform grid of neural features.
To query the density at a given position, the trilinear interpolated neural feature vector is passed through a small decoder network.
Given a set of X-ray input images, the neural features are optimized by differentiable volume rendering.
During the optimization, the octree structure is refined, and empty leaf nodes are removed from the tree.
Since all steps are differentiable, we can also perform self-calibration to compute the exact camera poses, the photometric detector response and the per-image capture energy.
An overview of our rendering pipeline is shown in Figure \ref{fig:pipeline_overview}.

\subsection{Image Formation}

The raw images of digital X-ray devices represent the transmission
images of a particular ray passing through an object.  The image
formation model in this setting derives from a continuous version of
the Beer-Lambert Law~\cite{kak2001principles}. For a given image pixel
$\pix$, the observed pixel value is given as
\begin{equation}
  \oimg(\pix) = \obgimg(\pix) \exp\left[-\int_{\ts}^{\te}
  \density(\ray_\pix(\ttt))\ d\ttt  \right],
    \label{eq:linspace}
\end{equation}
where $\obgimg$ is the (potentially spatially varying) intensity of
the x-ray source, $\ray_\pix$ is the ray associated with the image pixel
$\pix$, and $\ts$ and $\te$ are the ray parameters representing the
entry (near) and exit (far) point for a bounding box of the scene that
represents the reconstruction region. In tomographic imaging, we seek
to reconstruct the 3D distribution of $\density(\x)$ (the {\em
  attenuation cross section} or {\em density}) within this bounding
box.  This reconstruction is usually performed in logarithmic space,
i.e.
\begin{equation}
  \left(\bgimg(\pix) - \img(\pix)\right) =
  \int_{\ts}^{\te} \density(\ray_\pix(\ttt))\ d\ttt,
\label{eq:integral}
\end{equation}
with $\img= \log\oimg$ and $\bgimg=\log\obgimg$. In this formulation,
each pixel value $\img(\pix)$ is computed as the {\em line integral}
along the corresponding viewing ray $\ray_\pix$.
The discretization of \eqref{eq:integral} converts the integral into a finite sum.
\begin{equation}
(\bgimg(\pix)-\img(\pix)) \approx \sum _{i=1}^{N_p}\density(\ray_\pix(\ttt_i)) \delta_i.
\label{eq:sum}
\end{equation}
Here, $N_p$ represents the number of samples for the particular ray,
and $\delta_i$ denotes the length of the ray segment  covered by
sample $i$ (see next subsection).

\subsection{Ray Sampling}
\label{sec:sampling}
\begin{figure}
	\def\svgwidth{\columnwidth}
	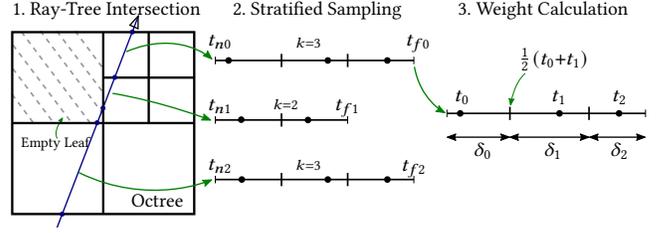
	\caption{
		Ray-sampling of the sparse octree structure. 
		(1.) Intersection intervals of non-empty leaf nodes are computed. 
		(2.) Samples are distributed by stratified random sampling. 
		Note here that the intervals $(\ts_0, \te_0)$ and $(\ts_2, \te_2)$ are assigned the same $k=3$ number of samples even though the latter covers more space.
		(3.) The integration weight $\delta$ is computed using Eq. \eqref{eq:weighting}.
	}
	\label{fig:tree_sampling}
\end{figure}

Given the octree structure and a specific ray $\ray$, the first step
in ray tracing is to generate a list of weighted samples
$\{(t_0,\delta_0), (t_1,\delta_1), \dots\}$.  This process is
visualized in Figure~\ref{fig:tree_sampling} and starts by computing
the ray segments $\{(n_0,\ts_0,\te_0), (n_1,\ts_1,\te_1), \dots\}$
that correspond to the intersection of the ray with octree node
$n_i$. Each segment consists of the node ID as well as the scalar ray
parameters $\ts_i, \te_i$ corresponding to the near and far points of
the segment. We then determine how many samples should be placed along
each of the segments as
\begin{equation}
k_i = \Bigl\lceil \ N \cdot \frac{\te_i-\ts_i}{diag(n_i)} \ \Bigr\rceil,
\end{equation}
with $N$ being a hyper parameter that represents the maximum number of
samples per node and $diag(n_i)$ being the diagonal size of node $n_i$.

This approach ensures $k_i \in [1,N]$ and that the number of samples is proportional to the relative length of the segment but independent of node size.
Small nodes obtain, on average, the same number of samples as large nodes, which in turn yields a higher sample density in subdivided regions that have been identified as regions of high geometric complexity.

Once $k_i$ has been determined, the interval is sampled either uniformly
(during the test stage) or with stratified random sampling (during the
training stage).  Finally, we compute the sample weight $\delta_j$ as:
\begin{equation}
\delta_j= 
\begin{cases}
    \frac{1}{2}(t_0+t_1) - \ts,& \text{if }j=0\\
    \te -  \frac{1}{2}(t_{k-2}+t_{k-1}) ,& \text{if }j=k-1\\
    \frac{1}{2}(t_j+t_{j+1}) - \frac{1}{2}(t_{j-1}+t_j),              & \text{else}
\end{cases}
\label{eq:weighting}
\end{equation}
Note that Eq. \eqref{eq:weighting} corresponds to numerical integration with central differences while related work, i.e. NeRF~\cite{mildenhall2020nerf}, make use of forward differences only. 
That approach is incompatible with hierarchical adaptive sampling because empty nodes in the middle break forward integration across node boundaries.

\subsection{Tree Query}
\label{sec:integration}
\begin{figure}
		\def\svgwidth{\columnwidth}
\begingroup%
  \makeatletter%
  \providecommand\color[2][]{%
    \errmessage{(Inkscape) Color is used for the text in Inkscape, but the package 'color.sty' is not loaded}%
    \renewcommand\color[2][]{}%
  }%
  \providecommand\transparent[1]{%
    \errmessage{(Inkscape) Transparency is used (non-zero) for the text in Inkscape, but the package 'transparent.sty' is not loaded}%
    \renewcommand\transparent[1]{}%
  }%
  \providecommand\rotatebox[2]{#2}%
  \newcommand*\fsize{\dimexpr\f@size pt\relax}%
  \newcommand*\lineheight[1]{\fontsize{\fsize}{#1\fsize}\selectfont}%
  \ifx\svgwidth\undefined%
    \setlength{\unitlength}{232.25943257bp}%
    \ifx\svgscale\undefined%
      \relax%
    \else%
      \setlength{\unitlength}{\unitlength * \real{\svgscale}}%
    \fi%
  \else%
    \setlength{\unitlength}{\svgwidth}%
  \fi%
  \global\let\svgwidth\undefined%
  \global\let\svgscale\undefined%
  \makeatother%
  \begin{picture}(1,0.31141897)%
    \lineheight{1}%
    \setlength\tabcolsep{0pt}%
    \put(0,0){\includegraphics[width=\unitlength,page=1]{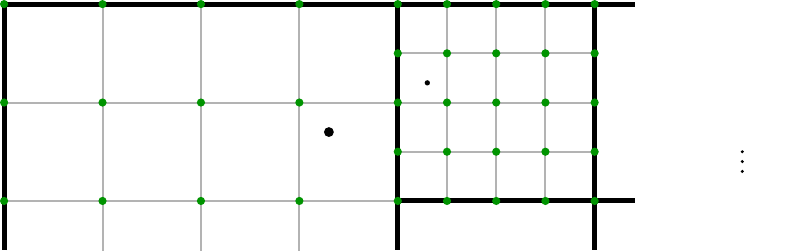}}%
    \put(0.83493451,0.28374045){\color[rgb]{0,0,0}\makebox(0,0)[lt]{\lineheight{1.25}\smash{\begin{tabular}[t]{l}$\scriptsize \text{Neural Features}$\end{tabular}}}}%
    \put(0,0){\includegraphics[width=\unitlength,page=2]{interpolation.pdf}}%
    \put(0.31216494,0.27856824){\color[rgb]{0,0,0}\makebox(0,0)[lt]{\lineheight{1.25}\smash{\begin{tabular}[t]{l}$\scriptsize \text{Non-Manifold}$\end{tabular}}}}%
    \put(0.52006082,0.01286172){\color[rgb]{0,0,0}\makebox(0,0)[lt]{\lineheight{1.25}\smash{\begin{tabular}[t]{l}$\scriptsize \text{Duplicate}$\end{tabular}}}}%
  \end{picture}%
\endgroup%

	\caption{
		Section of an octree structure that shows the alignment and sampling of the uniform grid inside each node.
		At the boundary of different nodes, some features are duplicated, and others are non-manifold.
	}
	\label{fig:interpolation}
\end{figure}

After sampling the rays, the next step is to retrieve the neural feature vector at the sample locations.
To that end, we compute the global coordinate $\x_g = \ray(\ttt)$ and convert it into local space of the containing leaf node.
This local coordinate is then used to sample a regular grid of neural features $\feat_i$ and interpolate with trilinear weights to obtain a feature vector $\feat(\x)$ for the sample location.
The process is visualized in Figure \ref{fig:interpolation}.
Note here that at the boundary between two nodes, duplicated and non-manifold features are stored in memory.
A regularizer is used to resolve this issue (see Section~\ref{sec:regularizer}).

Due to our sampling strategy (see Section~\ref{sec:sampling}), each ray and each node can have a different number of samples assigned to them.
To this end, we implement an indirect 3D grid sample kernel that can compute the neural feature vector from the local coordinate and the node ID in one step.
Experiments show that this custom layer is around three times more efficient than a sort-and-batch implementation using standard deep learning operators.

\subsection{Decoder Network}
\label{sec:decoder}

Once the feature vector has been obtained, we transform it into the desired output domain using a global decoder network
\begin{equation*}
  \mlp: \left\{\feat(\x)\right\} \rightarrow \density(\x)
\end{equation*}
that is shared among all nodes.
In the case of tomography, the output is a single scalar, representing the volume density $\density(\x)$ at the sample point $\x$.
The decoder itself is a three layer MLP with 64 neurons each, for a total of 4801 parameters to learn.
We use SiLU activation functions~\cite{elfwing2018sigmoid} inside the MLP and a single SoftPlus activation after the last layer to obtain a physically meaningful positive density value.
Finally, the density values are multiplied by the per-sample weight $\delta$ and summed up using a scatter-add operation resulting in the ray-integral of Eq. \eqref{eq:sum}.

\subsection{Loss and Regularization}
\label{sec:regularizer}

We optimize the neural features and the decoder's parameters using the mean squared error between the estimated and measured ray integrals, a total variation (TV) regularizer, and a boundary consistency (BC) regularizer.
\begin{equation}
\loss_{total} = \sum_\pix ||\img(\pix) - \img'(\pix) ||^2 + \lambda_{TV}\loss_{TV} + \lambda_{BC}\loss_{BC},
\end{equation}
where $\img'(p)$ is the estimated ray integral for ray $\ray_\pix$, $\img(\pix)$ the measured image intensity, and $\lambda_{TV}, \lambda_{BC}$ are hyper parameters that control the strength of each regularizer.

The TV loss acts as an additional spatial regularized during the
training, and has a similar role to the TV loss used in
optimization-based methods. To compute the TV loss, we utilize the
regular structure of the feature grids $\feat_n$ for each node. The
loss can either be computed directly on the feature vectors, or on the
decoded densities:
\begin{equation}
  \loss_{TV}^{(1)} = \sum_n\|\nabla\feat_n\|_1\quad\quad\text{or}\quad\quad
  \loss_{TV}^{(2)} = \sum_n\|\nabla\mlp(\feat_n)\|_1.
  \label{eq:tv}
\end{equation}
We experimented both variants and found that the former variant
produces slightly better results in addition to being faster.

The boundary consistency regularizer ensures a smooth transition between two neighboring nodes.
This is important, because as described in Section \ref{sec:integration}, duplicated and non-manifold features are stored along the node boundary.
This will result in block-like artifacts especially if only few images are used.
The regularizer minimizes the feature error on the boundary surface $\Lambda_{nm}$ for two neighboring octree nodes $n$ and $m$
\begin{equation}
\loss_{BC} = \sum_{(n,m)\in \mathcal{N}} \ \sum_{x \in \Lambda_{nm}} | \feat_m(x) - \feat_n(x)|,
\label{eq:bc}
\end{equation}
where $\mathcal{N}$ denotes the set of all pairs of neighboring nodes.

\subsection{Self-Calibration}
\label{sec:self_calibration}

CT reconstruction of real data often requires several calibration
steps both regarding the camera geometry and the radiometric
properties. 

\paragraph*{Radiometric self-calibration.}
As can be seen in Eq.~\eqref{eq:integral}, tomographic reconstruction
requires a reference image $\bgimg$ representing the illumination
pattern without an object present. In cone-beam CT, this image
captures effects such as the intensity dropoff towards the image
boundaries due to the cosine and $1/r^2$ terms, as well as any other
non-uniformities in the illumination. Unfortunately, adding or
removing the object from the setup can also disturb the validity the
reference image, causing artifacts in the reconstruction. The
differentiable nature of the NeAT framework allows us to refine (or
estimate from scratch) the reference image $\bgimg$, as well as a
per-view multiplier representing potentially different exposure times
for each view. Variations in exposure time are appropriate if the
object is much thicker in one direction than in another. Optimizing
these parameters can significantly improve the reconstruction quality,
depending on the dataset.

\paragraph*{Geometric self-calibration.}
High resolution tomographic reconstruction also relies on the
availability of high precision camera extrinsics and
intrinsics. Although in cone-beam CT the camera pose is usually
controlled with a high precision turntable, parameters such as the
precise field-of-view or the exact location of the rotation axis in
the image plane can be much harder to calibrate accurately, and in
fact they may drift over time due to heat expansion and other
factors. We instead propose to only use approximate parameter
estimation to get the camera model into the right ballpark, and to
then rely on gradient backpropagation to update the camera parameters,
including the relative positioning of the source and detector, as well
as the exact rotation angles between the views.


\subsection{Octree Update}
\label{sec:octree_update}
To automatically update the octree structure during the
reconstruction, we loosely follow the method of
ACORN~\cite{martel2021acorn}. However, while ACORN has access to a ground
truth image/volume data at every point in the cell, this ground truth
is not available in tomographic reconstruction tasks where the only
error metric available is the 2D reprojection error in each of the
views.

From this 2D image space error we estimate a volumetric error
distribution by summing over all the reprojection errors for ray
intersecting a given node. I.e. for a specific node $n$, the node
error becomes
\begin{equation}
  \err(n) = \density_{\max}(n) \cdot \sum_{(\pix,\ttt,\delta)\in\Omega_n} \delta || \img(\pix) - \tilde\img(\pix) ||^2,
\label{eq:node_error}
\end{equation}
where $\Omega_n$ refers to the set of all samples $(\ttt,\delta)$ in
$n$ from any ray passing through the node $n$, paired with the pixel
coordinates $\pix$ that generated the ray.  $\tilde\img$ refers to the
reprojection of the current volume estimate.  The summation term
therefore corresponds to a coarse tomographic reconstruction of the
reprojection error, integrated over the octree node. This volumetric
measure of error is additionally weighted by the maximum density of
the node, $\density_{\max}(n) =
\max_{(\ttt,\delta)\in\Omega_n}\density(\ray(\ttt))$.

Using this per-node error we then solve a mixed-integer program (MIP), which finds the best tree configuration with respect to a real-valued objective function.  Linear constraints ensure that at most $T_{max}$ leaf nodes are used and the configuration is a valid octree.
The details can be found in~\cite{martel2021acorn} as well as in our source code.


\section{Experiments}
%

\begin{figure*}
\includegraphics[width=\linewidth]{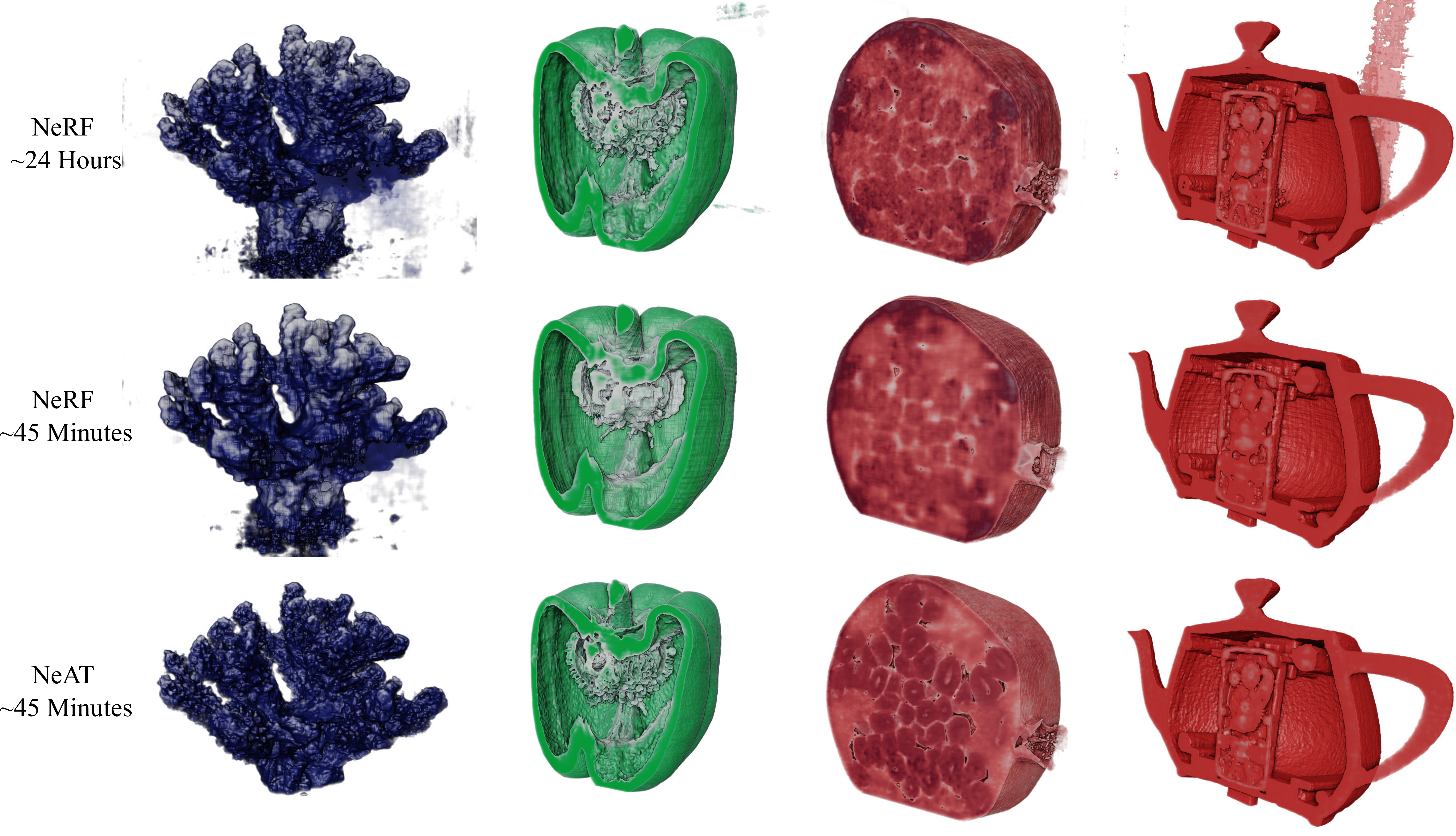}
  \caption{
  3D volume renderings of the real datasets, as reconstructed
  by NeRF and \neat using 25-50 projections. From left to right: ceramic coral, pepper, pomegranate, wind-up teapot. See Figure \ref{fig:real_recons} for a 2D slice comparison. }
  \label{fig:3D}
\end{figure*}

\subsection{Datasets and Evaluation Metric}
In the following, we present several reconstruction experiments on various CT datasets.
We divide these datasets into two classes: Real Data and Synthetic Data.
The real datasets are captured using a Nikon industrial CT scanner.
These images are noisy and some geometric and radiometric calibration errors are expected.
Since we don't have a ground-truth volume of the real datasets, we evaluate the performance using the reprojection error. 
In particular, the scanner provides us with a set of real X-ray images, which we split into a training and test set.
The training set is used to reconstruct the volume and the test set is used for the evaluation.
Figure~\ref{fig:3D} shows 3D renderings of \neat reconstructions for
all real datasets.

The synthetic datasets are given in the form of a volume sampled on a
regular voxel grid, which we use to generate synthetic X-ray images
from different angles.  By construction, there is no calibration error
or image noise in the rendered views, however we add some synthetic
Gaussian noise back in.  To evaluate the reconstruction quality of
synthetic datasets, we can directly compare the estimated volume
towards the ground-truth volume, although the reprojection error can
also be useful to assess overfitting.

\subsection{Ablation Studies}

\begin{figure}
\includegraphics[width=\linewidth]{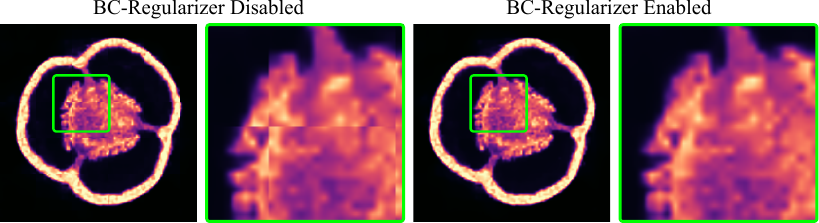}
	\caption{
		Sparse view tomography on the Pepper dataset with and without boundary consistency (BC) regularizer.
		Sharp edges at the boundary between neighboring octree nodes are successfully removed.
	}
	\label{fig:bc_reg}
\end{figure}

\begin{table}[]
\begin{tabular}{@{}l|lll|lll@{}}
\toprule
        & \multicolumn{3}{l|}{Sparse View} & \multicolumn{3}{l}{Limitted Angle} \\ \midrule
$\lambda_{TV}$        & Train   $\uparrow$        & Test    $\uparrow$        &       Vol. $\uparrow$   & Train       $\uparrow$      & Test   $\uparrow$      &       Vol. $\uparrow$       \\ \midrule
0.00000 & \textbf{43.97}  & 38.58     &  34.41   & \textbf{48.07}   & 26.37 & 22.20          \\
0.00002 & 43.21           & \textbf{39.32}  &  \textbf{35.96}  & 46.74            & 27.1 &22.81           \\
0.00010 & 41.74           & 39.17     &   35.25  & 44.95            & \textbf{27.92} & \textbf{23.56} \\
0.00025 & 40.69           & 38.63     &   34.67  & 43.50            & 27.36      & 23.38     \\
0.00050 & 39.44           & 37.47     &   33.10  & 42.67            & 26.89        & 23.02   \\ \bottomrule
\end{tabular}
\caption{ Reprojection error (PSNR) of the training and test views, as
  well as volumetric PSNR on the pepper dataset for different values of
  TV regularization.}
\label{tab:tvreg}
\end{table}

\paragraph{Regularziation} 
To regularize the reconstruction, we have implemented the TV- and BC-regularizer (see Section \ref{sec:regularizer}).
The BC-regularizer (Eq. \ref{eq:bc}) ensures a smooth transition between neighboring octree nodes. 
This is demonstrated in Figure \ref{fig:bc_reg}, which shows block-like artifacts if BC is disabled.
We found that $\lambda_{BC}=0.01$ gives good result on all datasets and is therefore used in the further experiments.
The TV-regularizer (Eq. \ref{eq:tv}) is used to further constrain underdetermined reconstruction problems, i.e., sparse view and limited angle tomography.
Table~\ref{tab:tvreg} shows both the reprojection errors on training views as
well as on previously unseen test views. In addition we show the
volume error directly.
The training and test reprojection errors demonstrate that overfitting can be reduced by increasing  $\lambda_{TV}$.
This also improves the  volumetric error.
We found that for sparse view problems $\lambda_{TV}=0.00002$ and for limited angle problems $\lambda_{TV}=0.00010$  gives the best results.

\paragraph{Geometric Self-Calibration}
\begin{figure}
\includegraphics[width=\linewidth]{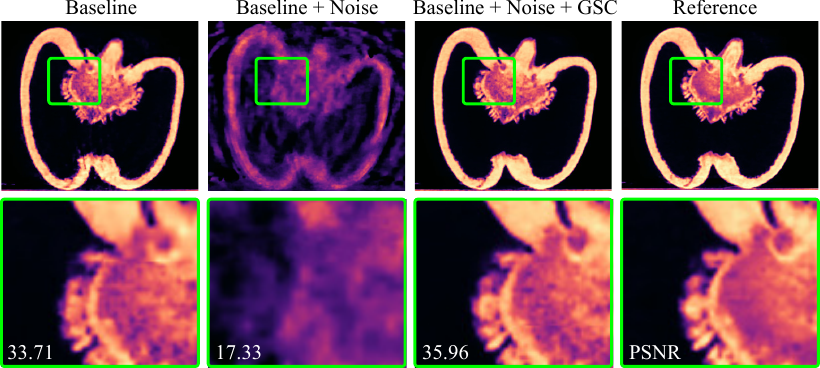}
	\caption{
		Ablation study of geometric self calibration on real data. 
		Baseline (left column) is the reconstruction using the initial calibration provided by the CT scanner.
		Adding noise to that calibration significantly degrades the result (second column).
		Our geometric self-calibration (GSC) can recover from the noisy input and even outperform the baseline calibration slightly (third column).
	}
	\label{fig:pose_opt}
\end{figure}
As described in Section~\ref{sec:self_calibration} our system can optimize the geometric parameters, such as detector orientation and source position, during the reconstruction.
In Figure \ref{fig:pose_opt}, we show the difference on a real dataset with geometric self-calibration enabled and disabled.
On the left, the baseline experiment is presented using the geometry configuration from the CT scanner.
Then we add Gaussian noise to the position and rotation of each view.
This degrades the reconstruction by 15 dB.
Using our geometric self calibration (GSC) our pipeline recovers from this bad initial calibration and then even outperform the baseline setting.
It is therefore robust and improves the reconstruction of real CT data.

\paragraph{Radiometric Self-Calibration}
\begin{figure}
\includegraphics[width=\linewidth]{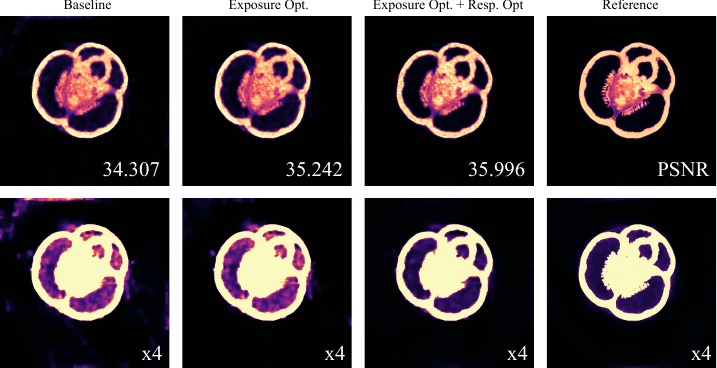}
	\caption{
		Sparse view tomography on the pepper dataset with and without radiometric self calibration.
		The bottom row shows the same volume multiplied by four to highlight noisy empty regions.	
	}
	\label{fig:photomeric}
\end{figure}
To test the effectiveness of radiometric self-calibration, we run our reconstruction pipeline on the real datasets and disables individual steps.
The results are presented in Figure \ref{fig:photomeric}.
The first experiment is the baseline with radiometric self calibration disabled.
After that we enable exposure and sensor bias estimation which improves the reconstruction by around 1 dB.
Enabling the response curve optimization, further improves the result, which can be seen in the amplified image at the bottom.

\paragraph{Decoder}
\begin{figure}
\includegraphics[width=\linewidth]{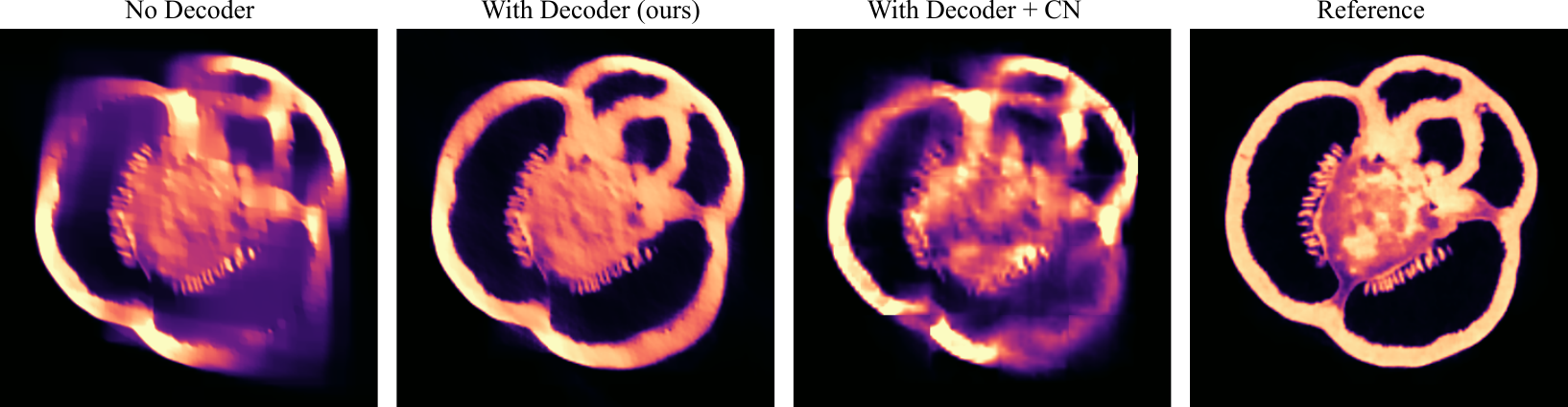}
	\caption{
		Limited angle reconstruction without decoder, with decoder, and with decoder and ACORN-style coordinate network (CN).
	}
	\label{fig:decoder}
\end{figure}
Next, we want to analyze the usefulness of the decoder network with respect to reconstruction quality and regularization properties.
In our current implementation, we use an eight-element feature vector which is transformed into a single density value by the decoder network.
The counterpart is a pipeline without decoder but double resolution blocks in x,y,z direction.
The volume has then exactly the same amount of variables but without the feature decoder.
Figure \ref{fig:decoder} shows the result of this comparison on a limited angle problem.
On the left hand side, we have the no-decoder variant with a grid resolution of $1\times33^3$. 
On the right, our full pipeline is displayed, which uses a grid resolution of $8\times17^3$. 
Using the decoder network, the sides of the fruit are reconstructed more accurately.
Without it, blurry artifacts appear, which are similar to artifacts of the iterative reconstructions methods.
We can therefore infer a regularizing property of the decoder network that improves the reconstruction in difficult conditions.

\begin{figure*}
\includegraphics[width=\linewidth]{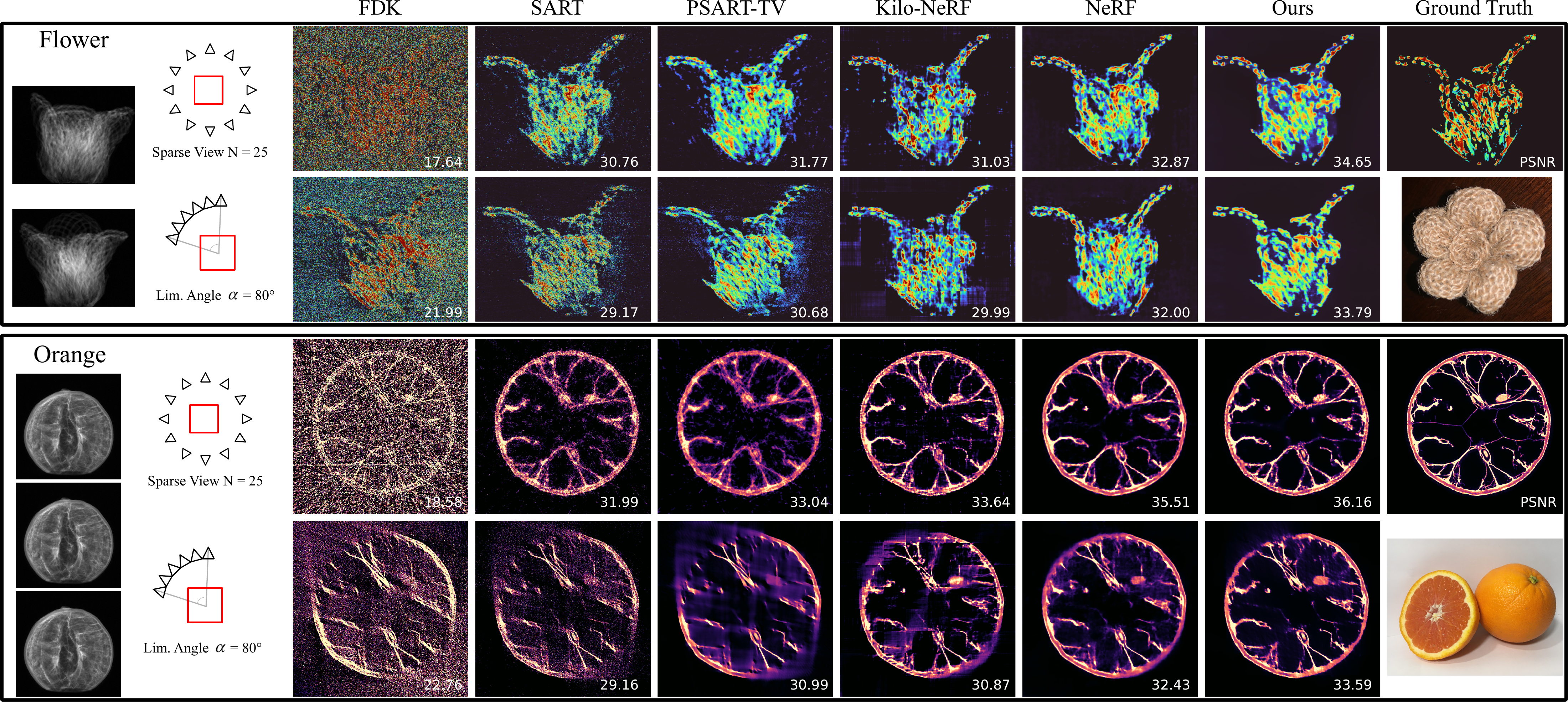}
	\caption{
		Sparse view and limited angle reconstructions on synthetic CT datasets.
		In the left most column, some raw input images are shown together with the reconstruction configuration.
		In the right most column, we show the Ground Truth, and illustrating images of the scanned objects.
		This comparison shows that our method (Ours) outperforms other baseline methods both qualitatively and quantitatively.
	}
	\label{fig:synth_recons}
\end{figure*}

\paragraph{Explicit vs. Hybrid explicit-implicit representation.}
The neural feature volumes in our reconstruction pipeline is stored
explicitly as a large tensor (see Figure \ref{fig:pipeline_overview}).
An alternative design would be a hybrid explicit-implicit model, like
ACORN~\cite{martel2021acorn} in which the feature volumes are not
stored explicitly, but are further compressed into an implicit neural
network in the hope of achieving additional compression and
regularization. Unfortunately, this hope does not materialize in the
context of tomographic reconstruction (Figure~\ref{fig:decoder}, third
sub-image). Specifically, we found that the PSNR for hybrid
explicit-implicit representations are worse than for our purely
explicit hierarchical representation, especially in the case of
limited angle tomography. In limited angle tomography, reconstructions
are already blurred in the direction orthogonal to the missing viewing
direction (missing wedge problem). The additional regularization of
the implicit network further encourages this blur instead of repairing
it.

Moreover, adding an implicit network dramatically increases the
training times. Finally, while final explicit-implicit representation
is more compact than our explicit one, the intermediate memory
consumption during training is actually higher, since all feature
volumes need to be decoded in order to trace all rays for one view.

\begin{figure}
\includegraphics[width=0.48\linewidth]{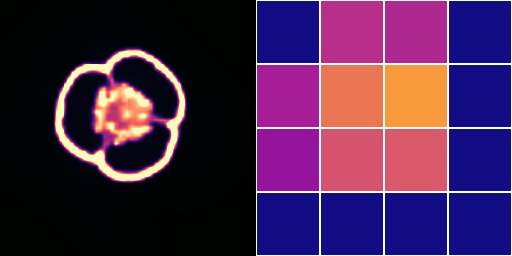}
\includegraphics[width=0.48\linewidth]{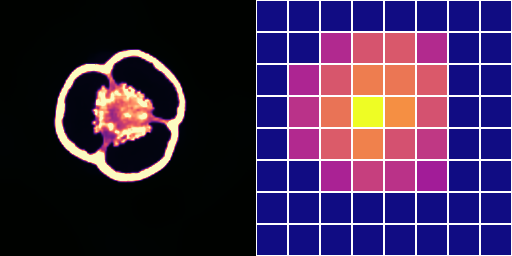}

\includegraphics[width=0.48\linewidth]{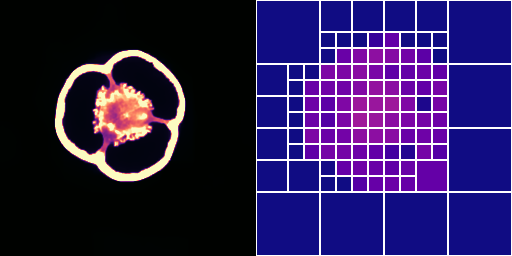}
\begin{tikzpicture}
		\begin{axis}[
			height=3cm,
			width=0.59\linewidth,
			at={(0,0)},
			ymin=0,
			ymax=0.01,
			xmin=0,
			xmax=6800,
			ymode=log,
			scaled x ticks = false,
			minor tick num =1,
			minor tick style={draw=none},
			minor grid style={thin,color=black!10},
			major grid style={thin,color=black!10},
					                     xlabel style={font=\tiny},
			ylabel style={font=\small},
			x tick label style={
				/pgf/number format/assume math mode, font=\tiny},
			y tick label style={
				/pgf/number format/assume math mode, font=\tiny},
			]
			\addplot[tension=1,color=green] table [x=step, y=mean, col sep=comma] {images/ablation_structure/loss.csv};
			\addplot[color=teal,domain=0:1, samples=500,dashed]{pow(x,1/2.2)};
		\end{axis}
\end{tikzpicture}

	\caption{
		Adaptive reconstruction of the pepper dataset.
		At the right side of each slice the octree structure is visualized.
		The color indicate the per-node error as defined in Eq. \eqref{eq:node_error}.
		This error is used to optimally distribute a fixed number of leaf nodes onto the scene.
		The bottom right shows the training loss during the reconstruction.
		The tree structure is optimized once a convergence has been detected, here at step 2200 and 5400.
	}
	\label{fig:structure}
\end{figure}
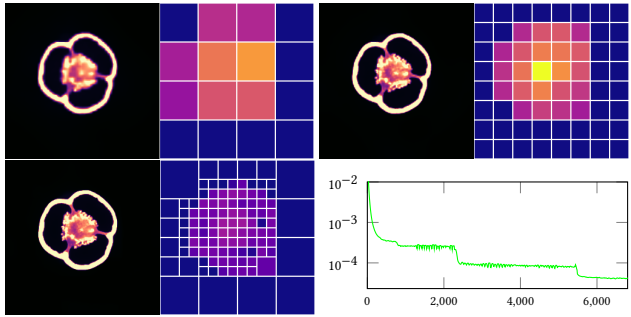

\paragraph{Structure Refinement}

In Section \ref{sec:octree_update}, we describe the structural octree optimization, which is performed every few epochs during the reconstruction.
This optimization consists of merging, splitting, and culling leaf nodes from the tree.
From these transformations, we expect a shorter reconstruction time, due to empty space skipping, and a more accurate reconstruction, due to allocating more resources to difficult parts of the scene.
The structure optimization process is presented in Figure \ref{fig:structure} on the pepper dataset with a maximum number of 1024 leaf nodes.
The octree is initialized by a uniform grid of resolution $4^3=64$.
Once the training has converged on that resolution the per-node error is computed using Eq. \eqref{eq:node_error} and the structure refinement is applied.
From top left to top right all leaf nodes are split because a full split doesn't exceed the leaf node budged $8^3 = 512 < 1024$.
However, at the next structure refinement step, splitting all nodes again is not possible $16^3 = 4096 > 1024$.
The optimization therefore automatically merges low-error nodes to split more high-error leaves.
Note that in this example, the empty space culling has been disabled to demonstrate adaptive node allocation.
The graph in the bottom left of Figure \ref{fig:structure} shows the loss curve during the reconstruction.
The steep steps indicate the points of structure optimization.

\subsection{Comparison to Existing Methods}

\begin{figure*}
\includegraphics[width=\linewidth]{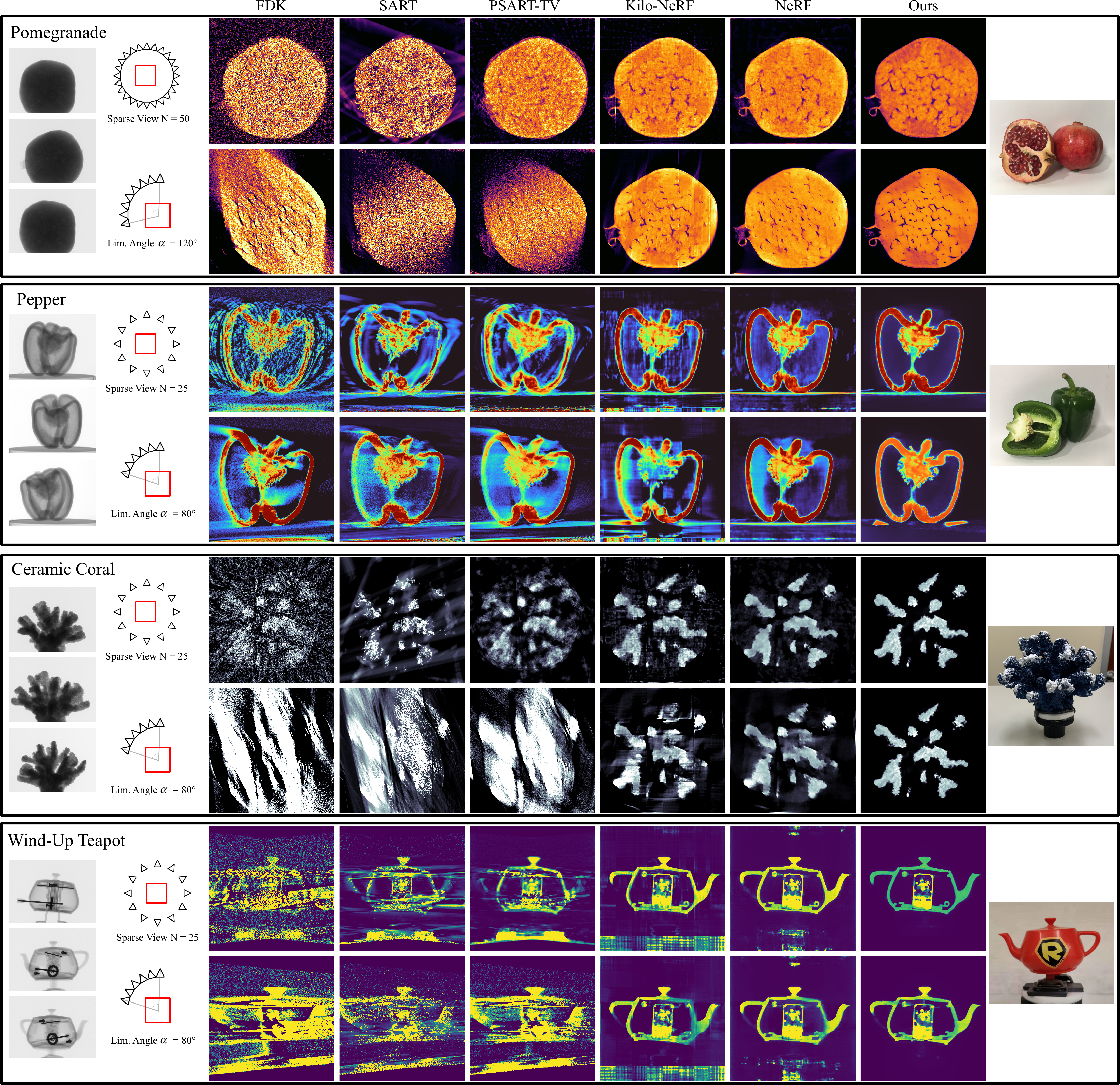}
	\caption{
		Sparse view and limited angle reconstructions on real CT datasets.
		In the left most column, some raw input images are shown together with the reconstruction configuration.
		In the right most column, we show illustrating images of the scanned objects.
		This comparison show that our method (Ours) achieves high-quality results for all data and for the two configurations.
	}
	\label{fig:real_recons}
\end{figure*}

We have evaluated our CT reconstruction method on various datasets and
compare it to other state-of-the-art approaches.  The other methods
are cone-beam filtered backprojection (FDK)~\cite{feldkamp1984practical}, 
the Simultaneous Algebraic Reconstruction
(SART)\cite{kak2001principles}, the proximal SART with TV regularizer
(PSART-TV)~\cite{zang2018super}, The Neural Radiance Fields (NeRF)
~\cite{mildenhall2020nerf}), and a NeRF variant making use of separate
local implicit functions (Kilo-NeRF) \cite{reiser2021kilonerf}.  The
first three are traditional CT-reconstruction approaches and variants
of these are usually found in commercial CT systems.  The latter two
are modern rendering approaches originally designed for novel view
synthesis and multi-view reconstruction. We have adapted them to
handle X-ray input data by changing the compositing operator to our
image formation model.  Furthermore, we modified Kilo-NeRF by
disabling the student-teacher distilling to validate if a direct
training of local MLPs is possible.

The algebraic reconstruction techniques are run until convergence
which takes between 5-30 minutes for simple methods like SART, according to the number of projections used, but
takes about 40-50 minutes on for more advanced methods like
PSART-TV. For the optimization-based methods, we conduct a parameter
search and present the best quality results.
NeRF and Kilo-Nerf
are trained for 24h and our approach is run for 40 epochs
corresponding to around 45 minutes on an A100 GPU.  When comparing the
times, it is important to keep in mind the vastly different code
bases, and the fact that all methods have a significant number of
parameters and hyper parameters that affect the timing. The provided
times are therefore only to be interpreted as rough indicators of
compute performance. Please also see the discussion in
Section~\ref{sec:discuss}.

In the first experiments, we measure the reconstruction quality on synthetic data.
The volume is known \textit{a priori} and the projection images are generated by volume rendering.
On each projection, we add Gaussian noise with a standard deviation of $\sigma = 0.02 \cdot I_{max}$.
No other calibrations errors are simulated.
The resulting reconstructions of each method are presented in Figure \ref{fig:synth_recons}.
The first two rows depict sparse-view and limited angle tomography on the flower dataset.
The same is done for the Orange dataset in the following rows.
All methods (except FDK) achieve decent results on both datasets and configurations.
However, on the Flower dataset our method is the clear favorite with a PSNR improvement of almost 2dB over the second best.
On the Orange dataset, both NeRF and Neat show a high quality reconstruction.
Quantitatively our method has the edge due to its sharpness but in the limited angle reconstruction (last row), NeRF is able to complete the orange's skin better.

After the synthetic experiment, we evaluate the same systems on four real datasets captured by a commercial CT scanner.
These datasets are Pomegranate, Pepper, Ceramic Coral, and Wind-Up Teapot.
They cover a wide range of interesting aspects such as the low-contrast internal structure of the pomegranate and the sophisticated mechanics of the wind-up teapot. 
Sample raw-images of each object are shown in Figure \ref{fig:real_recons} on the left.
Next to that, the current reconstruction configuration is visualized. 
On all four datasets and two configurations, NeAT provides the visually best reconstructions.
The volume is almost noise-free, edges are sharp and fine details such as the pepper seeds are preserved.
There are a few artifacts in the limited angle reconstruction, however the artifacts and noise of the other approaches are more severe.

\section{Discussion and Conclusions}
\label{sec:discuss}

\subsection{Discussion}
In the previous section we have compared \neat to other CT reconstruction approaches.
We have shown that our approach achieves significantly improved results than baseline methods both on synthetic and real datasets. We believe that the reasoning is multilateral.
First of all, we use a neural regularizer in the form of a decoder network that is able to steer the reconstruction to a physically plausible solution. 
Secondly, our geometric and photometric self calibration eliminates tiny errors of real-world data.
Lastly, the adaptive octree refinement ensures a smart distribution of memory and computational resources to complex parts of the scene.

In terms of compute times, \neat is comparable in compute time with
advanced optimization-based methods like PSART-TV, although precise
comparisons of compute time of course depend on the hyper parameters of
either method, as well as the degree of code optimization. 

Most of the computational effort of \neat is spent on evaluating the
decoder network, while the ray-tracing itself is substantially faster
than the PSART-TV implementation. This is primarily due to two factors:
our adaptive, octree-based approach that can efficiently allocate
samples to interesting volume regions, and our effective use of GPUs
as opposed to (already highly optimized) multicore CPU code in the
reference implementations of the optimization-based approaches.

It is therefore likely that a careful GPU implementation of a
hierarchical version of, for example, PSART-TV could realize similar
performance gains as \neat. However, doing so would be significantly
more difficult than for \neat: iterative solvers are based on a volume
projection operator $\mmat{A}$ and the corresponding backprojection
operator $\mmat{A}^T$. Since the matrices are far too large to store,
$\mmat{A}$ and $\mmat{A}^T$ are implemented procedurally as separate
operators, where $\mmat{A}$ is a gather operator while $\mmat{A}^T$ is
a scatter operator. Even with CPU code on a uniform grid it is not
trivial to get the two operators to perform well while maintaining an
exact transpose relationship to each other, which is a condition for
the convergence and correctness of most solvers. GPU implementations
with hierarchical data structures would dramatically complicate this
task further.

This is where the differentiable rendering approach of \neat shines:
instead of having to implement both operators, we only need to
implement the forward operator $\mmat{A}$ in a differentiable fashion,
and can then rely on backpropagation for the optimization. By using an
appropriate environment such as the PyTorch backend, the effort to
implement this efficiently on a GPU is much reduced.

\subsection{Limitations and Future Work}
Despite these advantages, during our experiments we also found some limitations that should be worked on in the future.
One general problem in adopting neural networks for scientific applications is that artifacts of traditional approaches are usually easy to spot for a human since they come in the form of strong blurriness or long streaks.
In the case of neural adaptive tomography, the artifacts look
physically plausible and are therefore harder to distinguish from a
correct reconstruction.
For example, in the teapot dataset (Figure~\ref{fig:real_recons}
bottom), all limited angle reconstructions exhibit a topology change
in the lower right corner. For the optimization-based methods it is
obvious that these regions cannot be trusted, whereas the deep
learning methods show plausible low-frequency completions of the
geometry, which however do not match reality.

As a further limitation, we also note that our method is currently
only suitable for a tomographic image formation model, which can be
evaluated in an order-independent fashion
\eqref{eq:integral}. Reconstructions of opaque objects require a
compositing image formation model similar to
NeRF~\cite{mildenhall2020nerf}, which requires all volume samples to
be ordered front to back. With our hierarchical octree subdivision of
space this would require additional book keeping efforts. Furthermore,
we believe the sampling strategy would likely have to be adapted for
this type of scene to concentrate the samples close to object
surfaces. 

\subsection{Conclusion}

In summary, we have presented \neat, the first neural rendering
architecture to {\em directly} train an adaptive, hierarchical neural
representation from images. This approach yields superior image
quality and training time compared to other recent neural rendering
methods. Compared to traditional optimization-based tomography
solvers, \neat shows better quality while matching the computational
performance.

While \neat is at the moment optimized for tomographic reconstruction,
we believe that similar concepts can be employed to reconstruct
complex scenes with opaque surfaces. This, however, we leave for
future work.


\begin{acks}

%
%
This work was supported by King Abdullah University of Science and
Technology as part of the VCC Competitive Funding as well as the CRG
program. The authors would like to thank Prof. Gilles Lubineau,
Dr. Ran Tao, Hassan Mahmoud, and Dr. Guangming Zang for helping with
the scanning of the objects.

\end{acks}

\bibliographystyle{ACM-Reference-Format}
\bibliography{common/10_Bibliography}

\end{document}